# Dirichlet Process Mixtures of Generalized Mallows Models

**Marina Meilă**
Department of Statistics
University of Washington
Seattle, WA 98195-4322
mmp@stat.washington.edu

**Harr Chen**
Computer Science and Artificial Intelligence Laboratory
Massachusetts Institute of Technology
Cambridge, MA 02139-4307
harr@csail.mit.edu

## Abstract

We present a Dirichlet process mixture model over discrete incomplete rankings and study two Gibbs sampling inference techniques for estimating posterior clusterings. The first approach uses a slice sampling subcomponent for estimating cluster parameters. The second approach marginalizes out several cluster parameters by taking advantage of approximations to the conditional posteriors. We empirically demonstrate (1) the effectiveness of this approximation for improving convergence, (2) the benefits of the Dirichlet process model over alternative clustering techniques for ranked data, and (3) the applicability of the approach to exploring large real-world ranking datasets.

## 1 MOTIVATION

Dirichlet process mixtures (DPM) are among the most successful ways of modeling multimodal distributions in a nonparametric Bayesian framework. They provide an elegant tradeoff between parameter sharing and parameter variability between clusters, are extremely versatile due to the flexibility in choosing base distributions, and enjoy all the other advantages of a fully generative probabilistic model. However, the feature that makes the DPM model so useful – the fact that it represents a fully nonparametric posterior – also poses its greatest challenge, in that the posterior is not computable in closed form. Therefore inference in a DPM must be performed using approximate techniques such as Monte Carlo sampling.

This paper introduces the DPM for the generalized Mallows (GM) model, a family of distributions over rankings. The GM has registered increasing popularity in recent years, partly because of a growing interest in ranked data and partly for its elegant computational properties [Lebanon and Mao, 2008, Meilă et al., 2007]. However, as an exponential family model, the GM is unimodal and thus suitable only for a limited range of applications when applied in isolation. By incorporating it in a model hierarchy like the DPM, we can enjoy the benefits of a fully generative multimodal model of ranked data.

To estimate the posterior of a DPM of GMs, we present two Gibbs sampling approaches. In the first, we explicitly draw all model parameters, relying on slice sampling for one of the posterior distributions. Our second approach marginalizes out several parameters by using approximations to the conditional posteriors, accelerating convergence at the cost of introducing error in the stationary distribution.

We conduct three sets of experiments exploring the properties of our approach. First, we compare the two sampling approaches and find that the approximation is beneficial for improving convergence. Second, we study the performance of the DPM of GMs in relation to previous clustering techniques for ranked data, showing improvements in held-out test likelihood. Third, we conduct a qualitative analysis of a large set of college admissions rankings, drawing conclusions that build upon observations made in previous work.

## 2 GENERALIZED MALLOWS MODELS

This section provides background on the generalized Mallows model, following Fligner and Verducci [1986]. Let $\pi$ denote a permutation over the set $\{1,2,3,\ldots,n\}$, where $\pi(l)$ is the rank of element $l$ in $\pi$ and $\pi^{-1}(j)$ is the element at rank $j$. One can uniquely determine any $\pi$ by the $n-1$ integers $V_1(\pi), V_2(\pi), \ldots, V_{n-1}(\pi)$ defined by $V_j(\pi) = \sum_{l>j} 1_{[l \prec_\pi j]}$. In words, $V_j$ is the number of elements in $\{j+1,\ldots,n\}$ that are ranked before $j$ by $\pi$. It follows that $V_j$ takes values in $\{0,\ldots,n-j\}$. Each element

$V_j$ can be set independently in specifying a $\pi$, which is not true of $\pi(l)$ values. These $V_j$'s are called the *code* of $\pi$.

This code can be defined w.r.t. to any *reference permutation* $\sigma$ by $V_j(\pi \mid \sigma) \equiv V_j(\pi\sigma^{-1})$. For any $\pi$ and $\sigma$ we define $s_1, \ldots, s_{n-1}$ to be a reciprocal form of the code, by exchanging the places of $\sigma$ and $\pi$:

$$s_j(\pi \mid \sigma) = V_j(\sigma \mid \pi) = \sum_{l \succ_\pi j} 1_{[l \prec_\sigma j]}. \quad (1)$$

Equivalently, $s_j$ is equal to one less than the rank of $\pi^{-1}(j)$ in $\sigma \setminus \pi^{-1}(1:j-1)$.

Based on this code representation, Fligner and Verducci [1986] introduced the following family of exponential models called the *generalized Mallows* (GM) models:

$$GM^s_{\vec{\theta},\sigma}(\pi) = \frac{e^{-\sum_{j=1}^{n-1} \theta_j s_j(\pi|\sigma)}}{\psi(\vec{\theta})}. \quad (2)$$

The $GM^s$ distribution is parametrized by the *central permutation* $\sigma$ and *concentration parameters* $\vec{\theta} \equiv \theta_{1:n-1} \geq 0$; $\psi(\vec{\theta})$ is a normalization constant that does not depend on $\sigma$:

$$\psi(\vec{\theta}) = \prod_{j=1}^{n-1} \psi_{n-j}(\theta_j) = \prod_{j=1}^{n-1} \frac{1 - e^{-(n-j+1)\theta_j}}{1 - e^{-\theta_j}}. \quad (3)$$

The $GM^s$ model factors into a product of independent univariate exponential models, one for each $s_j$:

$$P[s_j(\pi|\sigma) = k] = \frac{e^{-\theta_j k}}{\psi_{n-j}(\theta_j)}. \quad (4)$$

For $\theta_{1:n-1} = 0$, $GM^s$ is the uniform distribution. For $\theta_{1:n-1} > 0$, the $GM^s$ distribution has a unique maximum at $V_{1:n-1} = 0$, i.e., at $\pi = \sigma$. Thus the $GM^s$ is centered around $\sigma$ with exponential decay controlled by $\vec{\theta}$.

One can replace $s_j(\pi \mid \sigma)$ with $V_j(\pi \mid \sigma)$ in (2), obtaining a $GM^V$ with similar form to $GM^s$. These two models are equivalent only when all $\theta_j$ are equal.

### 2.1 TOP-$t$ RANKINGS

A permutation $\pi$ is a *top-t ranking* when one only observes the top $t$ ranks $(\pi^{-1}(1), \ldots, \pi^{-1}(t))$ rather than the entire permutation. In a top-$t$ ranking, the codes $s_{1:t}$ are fixed while the remaining $s_{t+1:n-1}$ are undetermined and can take any value in their respective ranges. For the $GM^s$ model the marginals w.r.t. $s_1, \ldots, s_t$ for some $t < n$ represent the probability of a top-$t$ ranking $(\pi^{-1}(1), \ldots \pi^{-1}(t))$ [Fligner and Verducci, 1986]. Meilă and Bao [2008] showed that the $GM^s$ model for top-$t$ rankings has sufficient statistics. In contrast, neither of these statements hold for $V_j$ codes and $GM^V$ over top-$t$ rankings.

In the rest of this paper, we will be considering data that consists of both full rankings and top-$t$ rankings of varying lengths (a full ranking is simply a top-$t$ ranking with $t = n - 1$). Thus our focus is on the $GM^s$ model, and GM should be understood to refer to $GM^s$.

### 2.2 SUFFICIENT STATISTICS AND CONJUGATE PRIOR

For a given permutation $\pi$ we define matrix $R_j(\pi)$ as

$$R_{j,ii'}(\pi) = 1_{[\pi^{-1}(j)=i \text{ and } i' \not\prec_\pi i]}, \quad (5)$$

and for a dataset $\pi_{1:N}$ of lengths $t_{1:N}$ we define $R_j(\pi_{1:N})$ as

$$R_j(\pi_{1:N}) = \sum_{k=1}^{N} R_j(\pi_k). \quad (6)$$

In words, each $R_j$ corresponds to a rank $j$, and element $R_{j,ii'}$ counts how many times $i$ was present at rank $j$, minus how many of those times $i'$ preceded $i$; $R_{j,ii} = 0$ for all $i, j$. If the data consists of top-$t$ rankings of different lengths, $R_j(\pi_{1:N})$ will depend only on those rankings of length at least $j$, and $R_j(\pi_{1:N}) = 0$ for $j > \max(t_{1:N})$. For datasets of varying lengths, we will refer to $\max(t_{1:N})$ as simply $t$.

For any top-$t$ ranking $\pi$ and complete ranking $\sigma$, we have $s_j(\pi \mid \sigma) = L_\sigma(R_j(\pi))$, where $L_\sigma(A)$ denotes the sum of the elements in the lower triangle of matrix $A$, after its rows and columns are permuted by $\sigma$ [Meilă and Bao, 2008].

Matrices $R_{1:t}(\pi_{1:N})$ are the sufficient statistics of the GM for both the central permutation $\sigma$ and the parameters $\vec{\theta}$ [Meilă and Bao, 2008]. The existence of finite sufficient statistics implies that the GM will have a conjugate prior, whose parameters are an equivalent sample size $\nu > 0$, and a set of equivalent sufficient statistics of the form $R^0_{1:t}$. This prior is fully described by Meilă and Bao [2008].

In many contexts, including our present clustering task, one desires to be uninformative w.r.t. to the central permutation while expressing knowledge about the parameters $\vec{\theta}$. In this case, the prior has the form

$$P^0(\sigma, \vec{\theta}; \nu, r) \propto e^{-\nu \sum_j [\theta_j r_j + \ln \psi_{n-j}(\theta_j)]}, \quad (7)$$

with $r = [r_1 \, r_2 \, \ldots \, r_t]$, $r_j > 0$ being a vector of positive parameters. This prior was used by Fligner and Verducci [1988]. The corresponding posterior is:

$$P(\sigma, \vec{\theta} \mid \nu, r, \pi_{1:N})$$
$$\propto e^{-\sum_j [(\nu r_j + L_\sigma(R_j(\pi_{1:N})))\theta_j + (\nu + N_j) \ln \psi_{n-j}(\theta_j)]}, \quad (8)$$

where $N_j$ is the number of data elements of length at least $j$. The priors presented here are defined up to a normalization constant. In general, there is no closed-form expression for this constant.

In summary, the GM is an exponential family model with simple sufficient statistics. Because the central permutation is an explicit parameter, this model is both more interpretable and tractable than other (exponential family) models over permutations. We use it as a building block for the Dirichlet process mixture model, which we briefly review below.

## 3 DIRICHLET PROCESS MIXTURE MODELS

A *Dirichlet process mixture* [DPM; Antoniak, 1974] is a generative clustering model. Generating data $\pi_{1:N}$ from a DPM of GMs involves these steps:

$$\begin{aligned} G &\sim DP(\alpha, P^0(\sigma, \vec{\theta} \mid \nu, r)), \\ \sigma_i, \vec{\theta}_i &\sim G, \\ \pi_i &\sim GM(\pi \mid \sigma_i, \vec{\theta}_i). \end{aligned}$$

First, a discrete distribution $G$ over GM distributions is sampled from the Dirichlet process prior. This prior takes as a parameter a distribution over $\sigma$ and $\vec{\theta}$, in our case the conjugate prior $P^0$. Next, a specific GM distribution with parameters $\sigma_i, \vec{\theta}_i$ is drawn from $G$. Data point $\pi_i$ is finally sampled from this GM distribution.

If we sample data sequentially from this model, then the $(N+1)^{th}$ sample will be distributed according to

$$\sigma_{N+1}, \vec{\theta}_{N+1} \sim \frac{1}{N+\alpha} \sum_{i \leq N} \delta_{\sigma_i, \vec{\theta}_i} + \frac{\alpha}{N+\alpha} P^0_{\nu,r}. \quad (9)$$

Hence, any finite sample will be a finite mixture of GMs, allowing the DPM to represent ranking data that is multimodal, with permutations clustered around several centers. Another characterization of the DPM is that each data point $\pi_i$ is associated with a cluster label $c_i \in 1, \ldots, C$, and each cluster $c$ with a set of GM parameters $\sigma_c$ and $\vec{\theta}_c$.

Unlike a finite mixture, the number of clusters in the DPM is itself a random variable. It will grow with the size of the data in a way controlled by the concentration parameter $\alpha$. This makes DPM models ideal for scenarios where the number of mixture components is not well-defined in advance. DPMs have found extensive practical applications in areas such as topic modeling [Teh et al., 2006], natural language processing [Liang et al., 2007], vision [Sudderth et al., 2005], and computational biology [Rasmussen et al., 2009].

Bayesian inference in the DPM model is typically conducted via MCMC [Neal, 2000] or variational methods [Blei and Jordan, 2006]. We focus on the former approach, where the goal is to produce samples drawn from the appropriate posterior distribution. In particular, if we are interested in parameter estimation, our objective is to draw samples from $P(c_{1:N}, \sigma_{1:C}, \vec{\theta}_{1:C} \mid \alpha, \nu, r, \pi_{1:N})$, where $c_i$ is the cluster assignment of data point $\pi_i$, and each cluster $c$ has GM parameters $(\sigma_c, \vec{\theta}_c)$.

While previous work [Neal, 2000] has made it straightforward to write the expression of this posterior (see the following sections), our main challenge is in making inference practical. Designing such methods and making them efficient for nontrivial model sizes $n$ and sample sizes $N$ is the main contribution of this paper.

---

**Algorithm** SLICE-GIBBS

**Input** Parameters $\nu, \alpha, t, r_{1:t}, T, T_{\text{Gibbs}}, T_{\text{Slices}}$, Data $\pi_{1:N}$ of lengths $t_{1:N}$

**Output** Samples $c_{1:N}, \sigma_c, \vec{\theta}_c$

Initialize $c_{1:N}, \sigma_c, \vec{\theta}_c$ randomly

Repeat $T$ times

1. *Resample cluster assignments*
   For all points $\pi_i$ sample $c_i$ according to

   $$P[c_i = c] \sim \begin{cases} \frac{N_{-i,c}}{N+\alpha-1} GM(\pi_i \mid \sigma_c, \vec{\theta}_c) & \text{if } N_{-i,c} \neq 0 \\ \frac{\alpha}{N+\alpha-1} \frac{(n-t_i)!}{n!} & \text{if } N_{-i,c} = 0 \end{cases}$$

   If $N_{-i,c} = 0$ for the sampled cluster, sample a new $\sigma_c \mid \pi_i$ and $\rho_c \mid \pi_i$ according to Step 2 below

2. *Resample cluster centers*
   For all clusters $c$, repeat $T_{\text{Gibbs}}$ times
   
   (a) Sample $\sigma_c$ by SAMPLE-$\sigma$-STAGEWISE
   (b) Sample $\vec{\theta}_c$ by SAMPLE-$\theta$-SLICE

Figure 1: SLICE-GIBBS algorithm for estimating a DPM of GMs.

## 4 THE SLICE-GIBBS SAMPLER

We first present a naïve Gibbs sampler for estimating a DPM of GMs, following the approach of Neal [2000]. Our main goal is to build a Gibbs Markov chain over cluster assignments $c_{1:N}$ whose stationary distribution is the desired model posterior. Taking advantage of exchangeability, we can sample each point $\pi_i$'s cluster assignment $c_i$ as if it were the last point to be generated, i.e., conditioned on the assignments of other data points. Assuming the cluster parameters $(\sigma_c, \vec{\theta}_c)$

---

**Algorithm** SAMPLE-$\sigma$-STAGEWISE

**Input** Parameters $\vec{\theta}$, sufficient statistics $R_{1:t}$, prior parameter $\nu$, optional prior parameters $R_{1:t}^0$

**Output** Sample $\sigma$

1. Calculate matrix $R = \sum_{j=1}^{t} \theta_j(R_j + \nu R_j^0)$, or $R = \sum_{j=1}^{t} \theta_j R_j$ if prior for $\sigma$ is uninformative

2. For $j = 1:n$ and while $R \neq 0$
   (a) Calculate column sums $\rho_{1:n}$ of $R$
   (b) Sample $\sigma^{-1}(j) = i$ w.p. $\propto e^{-\rho_i}$
   (c) Set row and column $\sigma^{-1}(j)$ of $R$ to zero

3. Fill in remaining ranks of $\sigma$ uniformly at random with items not yet selected

---

Figure 2: SAMPLE-$\sigma$-STAGEWISE algorithm for exactly sampling $\sigma$ from the conjugate posterior given $\vec{\theta}$.

---

**Algorithm** SAMPLE-$\theta$-SLICE

**Input** Parameters $\nu, t, r_{1:t}, T_{\text{Slices}}$, statistics $S_{1:t}(\sigma)$

**Output** Samples $\theta_{1:t}$

Initialize $\theta_{1:t}$ according to previous sample

For $j = 1:t$, repeat $T_{\text{Slices}}$ times

1. Sample $u \sim \text{Uniform}(0, \tilde{P}(\theta_j))$, where
$$\tilde{P}(\theta_j) = e^{-(\nu r_j + S_j(\sigma))\theta_j - (\nu + N_j)\ln \psi_{n-j}(\theta_j)} \quad (10)$$

2. Determine slice $[a, b]$ using step-out procedure

3. Repeatedly sample $\theta_j \sim \text{Uniform}(a, b)$ until $u < \tilde{P}(\theta_j)$, shrinking $[a, b]$ with rejected samples

---

Figure 3: SAMPLE-$\theta$-SLICE algorithm for slice sampling $\vec{\theta}$ given $\sigma$.

are known, this yields the following resampling update for the cluster assignment of data point $\pi_i$:

$$P(c_i = c \mid c_{-i}, \sigma, \vec{\theta}) \quad (11)$$
$$\propto \begin{cases} \frac{N_{-i,c}}{N+\alpha-1} GM(\pi_i \mid \sigma_c, \vec{\theta}_c) \\ \quad \text{if } N_{-i,c} \neq 0, \\ \frac{\alpha}{N+\alpha-1} \int GM(\pi_i \mid \sigma, \vec{\theta}) P^0(\sigma, \vec{\theta} \mid \nu, r) d\sigma d\vec{\theta} \\ \quad \text{if } N_{-i,c} = 0. \end{cases}$$

Here, $N_{-i,c}$ is the number of elements in cluster $c$, excluding data point $i$. In many applications of the DPM it is possible to integrate over cluster parameters and explicitly sample only cluster assignments (known as *collapsed* sampling). In the case of the GM, despite our use of a conjugate prior the marginalization over $\sigma$ and $\vec{\theta}$ is analytically intractable, in part because of the unknown normalization term. Thus for our first sampler we resort to building a Markov chain over the state space $(c_{1:N}, \sigma_C, \vec{\theta}_C)$, where each variable is explicitly resampled conditioned on the other variables. The algorithm is presented in Figure 1, while its steps are discussed in detail below.

To sample $c_i \mid \sigma, \vec{\theta}$ as in (11) we need to calculate the probabilities on the right hand side. This is straightforward for $N_{-i,c} > 0$, using (2). For $N_{-i,c} = 0$ we use the following Lemma (see Appendix for proofs).

**Lemma 1** *The marginal probability of a single observation is $P(\pi_i \mid \nu, r) = \frac{(n-t_i)!}{n!}$.*

Next we need $\sigma_c \mid \vec{\theta}_c, \pi_{i \in c}$. Let $R_j = R_j(\pi_{i \in c})$ be the sufficient statistics of cluster $c$, and $S_j(\sigma_c) = L_{\sigma_c}(R_j)$. These statistics are input to the algorithm described by Lemma 2.

**Lemma 2** *$P(\sigma \mid \vec{\theta}, \nu, r, \pi_{1:N})$ can be sampled exactly by Algorithm SAMPLE-$\sigma$-STAGEWISE (Figure 2).*

Sampling from $\vec{\theta}_c \mid \sigma_c, \pi_{i \in c}$ is more challenging. The main obstacle to straightforward sampling is the unknown normalization factor of this distribution. However, the posterior of each $\theta_{1:t}$ is independent and unimodal.[1] This suggests that *slice sampling* [Neal, 2003] is a viable way of drawing values for $\vec{\theta}_c$.

**Lemma 3** *$P(\vec{\theta} \mid \sigma, \nu, r, \pi_{i \in c})$ can be sampled using Algorithm SAMPLE-$\theta$-SLICE (Figure 3).*

The structure of Algorithm SAMPLE-$\theta$-SLICE follows directly from Neal [2000]. The full SLICE-GIBBS sampler, so named for its inclusion of a slice sampler, is presented in Figure 1. It alternates between resampling cluster assignments $c_i$ of data points and cluster parameters $\sigma_c$ and $\vec{\theta}_c$. Because the cluster parameters themselves form a Gibbs chain, we take $T_{\text{Gibbs}}$ steps to ensure convergence; furthermore, the slice sampler takes $T_{\text{Slices}}$ steps for each $\theta_j$ due to its serial correlation. In our experiments we find that $T_{\text{Gibbs}} = 10$ and $T_{\text{Slices}} = 3$ are typically sufficient values.

## 5 THE BETA-GIBBS SAMPLER

The previous section has demonstrated the difficulty of sampling from the conjugate posterior of a GM, and how it can be overcome by using slice sampling inside the Gibbs sampling step. We now present an alternative approach in which several sampling steps

---
[1]Only $\theta_{1:t}$ needs to be sampled, as $\theta_{t+1:n-1}$ does not affect the rest of the sampling procedure.

> **Algorithm** SAMPLE-$\sigma$-N1
>
> **Input** Top-$t$ ranking $\pi$, prior parameters $r_j, \nu$
>
> **Output** Sample $\sigma$
>
> 1. For $j = 1 : t$
>    (a) Sample $V_j = k$ w.p. $\propto Beta(\nu r_j + k, \nu + 2)$ for $k = 0 : n - j$
>    (b) Place $\pi(j)$ at the $(V_j + 1)^{th}$ previously unassigned position of $\sigma$
> 2. Fill the remaining ranks of $\sigma$ uniformly at random with items not in $\pi$

Figure 4: SAMPLE-$\sigma$-N1 algorithm for approximately sampling $\sigma$ from the conjugate posterior when $N = 1$.

and marginalizations will be done in closed form. The key insight is that the *infinite generalized Mallows model* [Meilă and Bao, 2008] can be used to *approximate* some of the sampling distributions.

The first result arises from the fact that as $n \to \infty$ the normalization constant $\psi_j$ approaches the value $\psi_\infty(\theta) = \frac{1}{1-e^\theta}$. This form of the normalization constant permits several computations in closed form.

**Lemma 4** [**Meilă and Bao, 2008**] *If the number of items $n$ is infinite and countable then:*

$$P(\theta_j \mid \sigma, \nu, r, \pi_{1:N}) =$$
$$Beta(e^{-\theta_j}; \nu r_j + S_j(\sigma), \nu + N_j + 1), \quad (12)$$
$$P(\sigma \mid \nu, r, \pi_{1:N}) \propto$$
$$\prod_{j=1}^t Beta(\nu r_j + S_j(\sigma), \nu + N_j + 1). \quad (13)$$

In the above, (12) uses the *Beta* distribution, and (13) uses the *Beta* function; $N_j$ is the number of rankings of length at least $j$ and $S_j(\sigma)$ is again $L_\sigma(R_j(\pi_{1:N}))$.

For the finite case, we define an analogue to the *Beta* function that arises in the marginalization of $\vec{\theta}$:

$$\tilde{Beta}(a, b, n) \equiv \int_0^\infty e^{-\theta a} \left( \frac{1 - e^{-(n+1)\theta}}{1 - e^{-\theta}} \right)^{-b+1} d\theta. \quad (14)$$

Using this representation it can be easily verified that for finite $n$,

$$P(\sigma \mid \nu, r, \pi_{1:N})$$
$$\propto \prod_{j=1}^t \tilde{Beta}(\nu r_j + S_j(\sigma), \nu + N_j + 1, n - j). \quad (15)$$

Note that as $n \to \infty$, $\tilde{Beta}(a, b, n) \to Beta(a, b)$, which will form the core of our approximation. We can now show the following Lemmas (see Appendix for proofs).

> **Algorithm** BETA-GIBBS
>
> **Input** Parameters $\nu, \alpha, t, r_{1:t}, T, T_{\text{Gibbs}}, T_{\text{Slices}}$, Data $\pi_{1:N}$ of lengths $t_{1:N}$
>
> **Output** Samples $c_{1:N}, \sigma_c, \vec{\theta}_c$
>
> Initialize $c_{1:N}, \sigma_c, \vec{\theta}_c$ randomly
>
> Repeat $T$ times
>
> 1. *Resample cluster assignments*
>    For all points $\pi_i$ sample $c_i$ according to
>
>    $$P[c_i = c] \sim$$
>    $$\begin{cases} \frac{N_{-i,c}}{N+\alpha-1} \prod_{j=1}^t \frac{Beta(s_j(\pi_i|\sigma_c) + \nu r_j + S_j(\sigma_c), \nu + N_{c,j} + 2)}{Beta(\nu r_j + S_j(\sigma_c), \nu + N_{c,j} + 1)} \\ \quad \text{if } N_{-i,c} \neq 0 \\ \frac{\alpha}{N+\alpha-1} \frac{(n-t_i)!}{n!} \quad \text{if } N_{-i,c} = 0 \end{cases}$$
>
>    If $N_{-i,c} = 0$ for the sampled cluster, sample a new $\sigma_c|\pi_i$ by SAMPLE-$\sigma$-N1
>
> 2. *Resample cluster centers*
>    For all clusters $c$, if $N_c > 1$ repeat $T_{\text{Gibbs}}$ times
>    (a) Sample $\sigma_c \mid \vec{\theta}_c$ by SAMPLE-$\sigma$-STAGEWISE
>    (b) Sample $\vec{\theta}_c \mid \sigma_c$ using (12) from Lemma 4
>
>    If $N_c = 1$ sample $\sigma_c$ by SAMPLE-$\sigma$-N1

Figure 5: BETA-GIBBS algorithm for estimating a DPM of GMs.

**Lemma 5**

$$\sum_{s=0}^n \tilde{Beta}(s+a, N+1, n) = \tilde{Beta}(a, N, n). \quad (16)$$

*This result also holds as $n \to \infty$.*

**Lemma 6** *Marginalizing over $\vec{\theta}$ for a single $\pi$ yields:*

$$P(\pi \mid \sigma, \nu, r, \pi_{i \in c}) = \quad (17)$$
$$\prod_{j=1}^t \frac{\tilde{Beta}(s_j(\pi \mid \sigma) + \nu r_j + S_j(\sigma), \nu + N_j + 2, n - j)}{\tilde{Beta}(\nu r_j + S_j(\sigma), \nu + N_j + 1, n - j)}.$$

**Lemma 7** $P(\sigma \mid \nu, r, \pi)$ *(i.e., when $N = 1$) can be sampled approximately by Algorithm* SAMPLE-$\sigma$-N1 *(Figure 4).*

Lemmas 6 and 7, together with Lemmas 1 and 2, allow us to (approximately) marginalize out the continuous $\vec{\theta}$ parameters for much of the sampling. This algorithm, called BETA-GIBBS because of the extensive use of the *Beta* function, is given in Figure 5. The algorithm approximates $\tilde{Beta}(a, b, n)$ with the easily computable $Beta(a, b)$ for sampling $P(\vec{\theta}_c \mid \sigma_c)$, $P(c_i \mid \sigma_c)$ when $N_{-i,c} > 0$, and $P(\sigma_c \mid \pi)$ when $N_c = 1$. For the resampling of $\sigma_c$ for non-singleton clusters, we resort to an inner Gibbs sampler, setting $T_{\text{Gibbs}}$ to 10 as with SLICE-GIBBS.

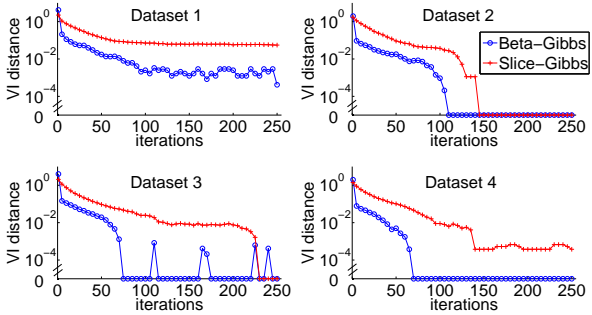

Figure 6: Performance of SLICE-GIBBS and BETA-GIBBS on four artificial datasets, averaged over ten replicates. Each plot displays VI distance to the true data labeling. Lower is better.

## 6 EMPIRICAL COMPARISON OF SLICE-GIBBS AND BETA-GIBBS

The purpose of introducing the BETA-GIBBS sampler was (1) to make the resampling of the parameters more efficient, and (2) more importantly, to reduce variance and accelerate convergence to the stationary distribution, which is a typical effect of marginalizing over certain parameters. We now verify how well we succeeded by running experiments on four artificial datasets under varying conditions. For each experiment, we generate 500 points from each of 10 clusters for a total of $N = 5000$ samples. Each cluster's points are generated from a GM with true $\sigma^*$ and $\vec{\theta}^*$ given below. To ensure that the dataset is not too easily separable, each $\sigma^*$ is drawn from the conjugate posterior of $\sigma$, conditioned on 100 permutations drawn randomly from a GM with $\vec{\theta} = 0.7$.

| Dataset | $n$ | $t$ | $\vec{\theta}^*$ |
|---|---|---|---|
| 1 | 20 | 10 | $\theta_i^* = 1$ |
| 2 | 20 | 19 | $\theta_i^* = 1$ |
| 3 | 20 | 10 | $\theta_i^* = 1.5 - (i-1) \times 0.1$ |
| 4 | 20 | 19 | $\theta_i^* = 1.5 - (i-1) \times 0.05$ |

We measure the Variation of Information (VI) distance [Meilă, 2007] between the sampled and true clusterings at each iteration. We average over ten runs for each dataset, initializing randomly with 20 clusters. Priors $\alpha$, $\nu$, and $r_{1:t}$ are all set to one.

Figure 6 shows the results of this experiment. In every case, BETA-GIBBS converges to the true clustering much more rapidly than SLICE-GIBBS. Furthermore we note that each iteration of SLICE-GIBBS is typically slower than BETA-GIBBS, due to the additional iterations of slice sampling. Interestingly, the comparison does not change when complete rankings are observed (datasets 2 and 4), where the $Beta$ approximation of $\tilde{Beta}$ is poorest. These results support the use of the BETA-GIBBS approximation for estimating a DPM of GMs in the general case.

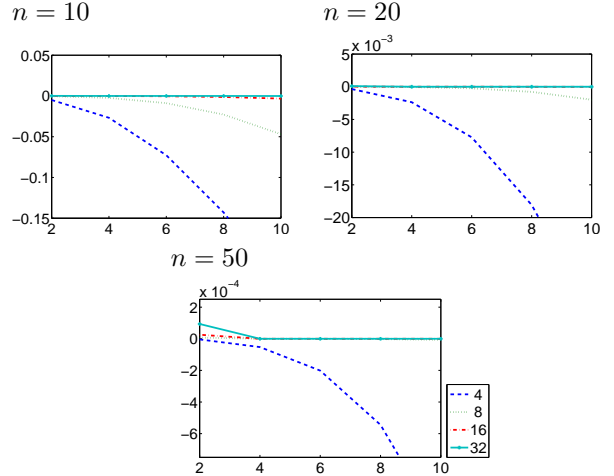

Figure 7: Relative error $(1 - Beta(a,b)/\tilde{Beta}(a,b,n))$ as a function of $a$, the equivalent number of inversions, for various values of $b$, the equivalent sample size plus two, for $n = 10, 20, 50$.

We also assessed the quality of our $Beta(a,b)$ approximations w.r.t. the correct values $\tilde{Beta}(a,b,n)$ in Figure 7. The approximation will be more accurate for larger $n$, so we consider the values $n = 10, 20, 50$ and $\nu = 1$. The relative error is largest for small $b$ and large $a$. This would occur in very small clusters, and worsens when consensus worsens, when the parameter prior is more diffuse (i.e., with large $r_j$), and when $j$ is higher. The effect is to overestimate the $\theta_j$'s for higher ranks and the probability of assigning points to small clusters.

## 7 COMPARISON TO RELATED WORK

Modeling of multimodal ranking data has been attempted in a variety of paradigms. For instance, top-$t$ ranked data representing votes in Irish elections has been modeled as mixtures of Plackett-Luce and Benter models by Gormley and Murphy [2008a] and Gormley and Murphy [2008b], respectively. Busse et al. [2007] developed an EM algorithm for estimating finite mixtures of GM models for top-$t$ rankings.

Of the nonparametric methods, the most flexible and theoretically principled is the Kernel Density Estimator (KDE) of Lebanon and Mao [2008], in which the kernel is the GM model with $\theta_j = \theta$, and where the data is partial rankings of a given type. One of the main algorithmic contributions of Lebanon and Mao is the tractable evaluation of the kernel, which includes summation over entire cosets of super-exponential cardinalities. Meilă and Bao [2008] introduced the exponential blurring mean-shift (EBMS) clustering algorithm, which also uses the GM model with all-equal $\theta_j$'s as the kernel. The algorithm features a heuristic method for estimating the kernel width $\theta$ and does not need a stopping rule, so it requires no outside parameters. Finally, Guiver and Snelson [2009] present an example of elegant Bayesian estimation where intractable inference in the Plackett-Luce model is approximated efficiently.

The most relevant comparisons to this work are with the nonparametric approaches of EBMS [Meilă and Bao, 2008] and KDE [Lebanon and Mao, 2008]. The former is not a generative model, and has various parametric limitations: the kernel width $\theta$ is represented by a single parameter and the output rankings are truncated at some user-set standard length. The latter model is generative, and applies to any type of partial rankings, which is beyond the scope of our current model. On the other hand, KDE has only one parameter for kernel width $\theta$, just like EBMS, and this parameter must be manually set. This is a limitation since in many instances of real-world data higher ranks are more concentrated around the mean than lower ranks.

### 7.1 EXPERIMENTS

We now conduct experimental comparisons of DPM, EBMS, and KDE on both artificial and real data. For the former, the data is generated from a fixed mixture model with $K = 3$ mixture components, all having the same single parameter $\theta_{1:t} = 1$.[2] We set $n = 12$, $t = 5$, and varied $N$ from 100 to 10000. We fit each of the three models to this data and calculate log-likelihood on a held-out test set. For DPM and EBMS, we also calculate the quality of the obtained clustering using VI distance [Meilă, 2007]. To give an idea of the best achievable performance, we use the same criteria to evaluate the true model. For KDE the kernel width is set to the true $\theta$. These conditions are the most favorable for the competing alternatives to the DPM.[3]

---

[2] An experiment with $K = 30$ clusters produces similar results, though then EBMS does not scale well to large $n$.

[3] The chosen kernel width is not provably optimal for KDE, as the optimal kernel width varies with the sample size. However, we have tested the KDE model under a wide range of sample sizes, and it is very likely that $\theta = 1$ is near optimal for at least one of these.

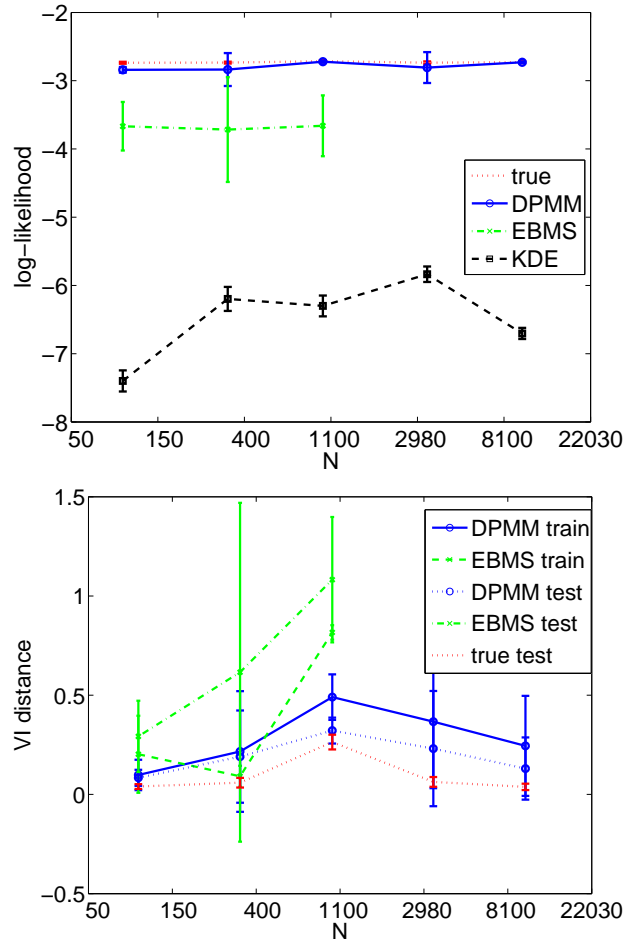

Figure 8: Performance of DPM, EBMS, and KDE on a mixture of $K = 3$ Mallows models, with $n = 12$, $t = 5$, and training sample sizes $N = 100, \ldots, 10000$, averaged over 10 replicates. The test set size is 3000. Top: test set log-likelihood, higher is better; bottom: VI distance to true data labeling, lower is better. EBMS was too slow for the larger $N$'s.

The results are shown in Figure 8. One sees that DPM, even though it has more parameters than necessary to explain the data, clearly performs better than EBMS and KDE in terms of likelihood, being almost equal to the true model. The $\sigma$ and $\theta$ estimates are also centered on the true values (not shown). On VI distance, the heuristic EBMS performs surprisingly well on the training data, occasionally surpassing DPM, but produces poor clusters on the test data. The VI of the true model shows that this is a case of well-separated mixtures, but not a trivial one.

We run a similar comparison on the `Jester` dataset [Goldberg et al., 2001], which consists of joke preferences for a large group of people. Of the 100 available jokes we restrict the data to the $n = 70$ most frequently rated, and to the top $t = 5$ rankings. As Figure 9 shows, we again observe that DPM outperforms KDE over all kernel widths tried. DPM finds between 4 and 9 clusters in 10 trials, with $\theta_j$ in the range $(0.03, 0.06)$ for all $j$'s and $N$'s.

| Rank | Cluster 1 (8.1%) | Cluster 2 (6.2%) | Cluster 3 (6.0%) | Cluster 4 (5.8%) | Cluster 5 (5.7%) |
|---|---|---|---|---|---|
| 1 | Business (Dublin) | Comp Appl (Dublin) | Business (Limerick) | Engineering (Dublin) | Comp Appl (Dublin) |
| 2 | Commerce (Dublin) | Comp Sys (Limerick) | Commerce (Galway) | Engineering (Galway) | Engineering (Dublin) |
| 3 | Business (Dublin) | Software Dev (Limerick) | Commerce (Cork) | Civil Eng (Galway) | Engineering (Dublin) |
| 4 | Marketing (Dublin) | Comp Appl (Cork) | Business (Waterford) | Engineering (Dublin) | Comp Sci (Dublin) |
| 5 | Marketing (Dublin) | Appl Comp (Waterford) | Humanities (Galway) | Elec Eng (Limerick) | Comp Sci (Dublin) |
| 6 | Humanities (Dublin) | Comp Network (Carlow) | Humanities (Cork) | Mech Eng (Limerick) | Science (Dublin) |
| 7 | Bus & Econ (Dublin) | Software Dev (Cork) | Admin (Limerick) | Engineering (Dublin) | Engineering (Dublin) |
| 8 | Bus & Legal (Dublin) | Software (Athlone) | Business (Dublin) | Elec Eng (Galway) | Comp Sci (Dublin) |
| 9 | Acct & Fin (Dublin) | Comp Sci (Maynooth) | Business (Dublin) | Info Tech (Limerick) | Computing (Dublin) |
| 10 | Acct & HR (Dublin) | Info Tech (Limerick) | Comp Sys (Limerick) | Civil Eng (Cork) | Comp Sci (Maynooth) |
| | **Business, Dublin** | **Comp Sci, ex-Dublin** | **Business, ex-Dublin** | **Engineering** | **Comp Sci, Dublin** |

Table 1: Top courses of the five largest clusters found in a representative run of the DPM over college admissions data. Proportions of the data assigned to each cluster is shown next to the cluster number. We list course names and school locations, and summarize the theme of each cluster.

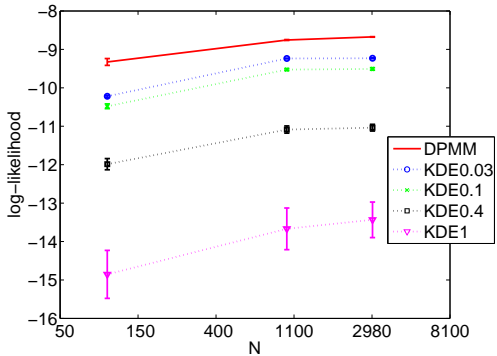

Figure 9: Test set log-likelihood of DPM and KDE on the `Jester` data, with $n = 70$, $t = 5$ and training sample sizes $N = 100, 1000, 3000$, averaged over 10 replicates. Higher is better. The test set size is 3000. Further reducing the kernel width for KDE leads to results almost identical to the width 0.03 case.

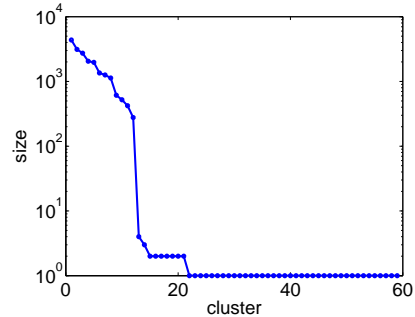

Figure 10: Cluster size distribution for a representative run of the college applications data, $N = 53757$. Out of 177 clusters, 114 were singletons, 35 had at least 54 points (0.1%), and 27 had at least 538 points (1%).

## 8 ANALYSIS OF COLLEGE COURSE RANKINGS

We also conduct an analysis of Irish third-level college applications, where prospective students rank up to ten preferred academic courses across a number of schools [Gormley and Murphy, 2006]. In combination with examination scores, this data is used by the Central Applications Office (http://www.cao.ie/) to determine placements into third-level degree programs.

The dataset consists of $N = 53757$ students in the year 2000 selecting from $n = 533$ courses and ranking up to $t = 10$ of them. To facilitate a comparison with previous work [Gormley and Murphy, 2006], we set $\alpha$ and $\nu$ to 100 so as to induce a finer-grained cluster-ing, running four samplers to 500 iterations each. The four runs yield between 23 and 27 substantial clusters (those with at least 1% of the data points), with similar central permutations recurring across runs. Figure 10 illustrates the sizes of the induced clusters.

Table 1 displays the top-10 courses from the central permutations of the largest clusters of a representative run. The results show clear thematic consistency in the top ranked courses by vocation and/or location, concordant in particular with Gormley and Murphy's observation of the "frequent distinction between sets of applicants who apply for degrees of a similar discipline but are deemed separate on the basis of whether or not the institutions to which they apply are in Dublin." Notably, their analysis revealed distinct clusters of computer science preferences, one for Dublin-based schools and one with regional variation. We additionally find a clear separation between Dublin-based business programs and outside business programs, a phenomenon that was observed by Gormley and Murphy but not explicitly identified by their clustering.

We can also interpret the posterior samples of $\vec{\theta}$ to gain insight into data separation by rank, which is an advantage of using the GM for modeling clusters compared to the Plackett-Luce model. We compute an average of each $\theta_j$, weighted by cluster size, across the four runs. We also perform this analysis for only large and only small clusters, thresholding at a size of 5% of the data points (splitting the data points roughly equally into large and small). Figure 11 presents these averages. The clearly decreasing overall trend reinforces the intuition that top-ranked choices tend to be more coherent and distinctive than later entries in the top-10 ranking. Furthermore, we find that small clusters tend to diverge less at top ranks than large clusters, but this trend reverses around the fourth rank. A qualitative examination of the data suggests that this may be because small clusters tend to correspond to more specialized interests with fewer relevant courses (e.g., courses at one specific smaller school), leaving fewer choices for the top ranks but allowing for greater divergence later on.[4]

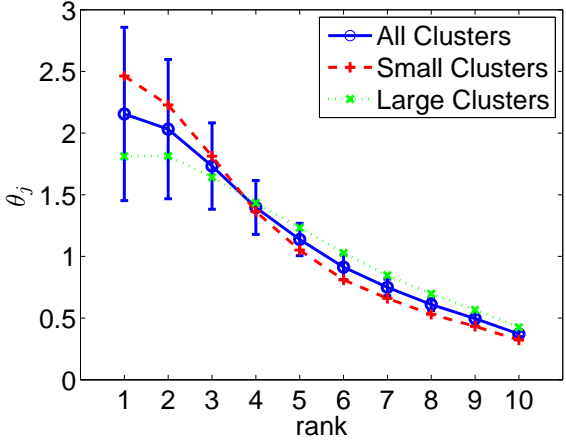

Figure 11: Average of $\theta_j$ weighted by cluster size as a function of rank $j$ for the college application data, replicated across four runs. The decreasing trend supports the intuition that top course preferences are better separated than less-desired choices.

## 9 DISCUSSION

We introduced nonparametric Bayesian DPMs on ranked data domains with top-$t$ rankings of variable lengths. Our inference algorithms are able to run on substantial dataset sizes and large $n$'s.

We leveraged a combination of statistical and computational insights in developing our techniques. Statistically, the rich combinatorial structure of the parameter space allowed us to perform explicit marginalizations and normalizations in special cases. Computationally, we exploited the special structure of the $R_j$ sufficient statistics and of the $L_\sigma$ operator, thereby eliminating $n$ from the most intensive computations. While the faster BETA-GIBBS algorithm uses approximate posteriors, we have verified empirically the quality of that approximation and the advantages it yields to convergence.

Our algorithm works with informative priors as well, with only a minor modification (replacing SAMPLE-$\sigma$-N1). One avenue of future work is to explore other sampling schemes than the one described by Neal [2000], such as split-merge algorithms [Jain and Neal, 2007].

## Acknowledgments

Harr Chen is supported by a National Science Foundation Graduate Fellowship.

## Appendix

**Proof of Lemma 1** The marginal of a single $\pi$ of length $t$ is

$$Z_1(\pi) = \sum_\sigma \int_0^\infty GM^s(\pi_i; \sigma, \vec{\theta}) P^0(\sigma, \vec{\theta}; \nu, r) d\vec{\theta}. \tag{18}$$

By Lemma 6, the integral is equal to

$$\frac{1}{n!} \prod_{j=1}^t \frac{\tilde{B}eta(\nu r_j + s_j(\pi \mid \sigma), \nu + 2, n - j)}{\tilde{B}eta(\nu r_j, \nu + 1, n - j)}. \tag{19}$$

Note that $s_j(\pi \mid \sigma) = V_j(\sigma \mid \pi)$, where $V_j(\sigma \mid \pi)$ should be read as "the rank in $\sigma$ of item $j$ of $\pi$" and is therefore well defined for $j = 1 : t$.

Any configuration of $V_j$'s uniquely determines a subset of the positions in $\sigma$, and the $V_j$'s can take any value in their admissible range when $\sigma$ ranges over all infinite permutations. Thus, $s_j(\pi \mid \sigma)$ ranges from 0 to $n - j$, and consequently the summation over $\sigma$ commutes with the product over $j$. For every configuration of $s_{1:t}$, there will be $(n-t)!$ different permutations with that configuration. It follows that:

$$\sum_\sigma \prod_{j=1}^t \tilde{B}eta(\nu r_j + s_j(\pi \mid \sigma), \nu + 2, n - j)$$
$$= (n-t)! \prod_{j=1}^t \sum_{s_j=0}^{n-j} \tilde{B}eta(\nu r_j + s_j, \nu + 2, n - j)$$
$$= (n-t)! \prod_{j=1}^t \tilde{B}eta(\nu r_j, \nu + 1, n - j).$$

---

[4]In fact, the average ranking length $t$ for data points in small clusters is shorter than for large clusters: 6.15 compared to 6.63.

The last equality is obtained from Lemma 5. Hence, $Z_1(\pi) = (n-t)!/n!$.

**Proof of Lemma 2** From Meilă and Bao [2008], for any given $\theta$,

$$\begin{aligned} P(\sigma \mid \theta, \pi_{1:N}, \nu, r) & \\ \propto\ & e^{-\sum_{j=1}^{t}[\theta_j(L_\sigma(R_j(\pi_{1:N}))+\nu r_j)+(N+\nu)\ln\psi_{n-j}(\theta_j)]} \\ \propto\ & e^{-\sum_{j=1}^{t}\theta_j L_\sigma(R_j(\pi_{1:n}))} = e^{-L_\sigma(R)}. \end{aligned}$$

We use a key observation of Meilă et al. [2007], which is that for a distribution over permutations like the one above, the first rank of $\sigma$ is distributed proportionally to the column sums of $R$, the second rank is distributed proportionally to the column sums of $R$ after deleting row and column $\sigma^{-1}(1)$, etc. Hence, the ranks of $\sigma$ can be sampled sequentially by

$$P(\sigma^{-1}(1) = k) \propto e^{-\sum_i R_{ik}}, \quad (20)$$
$$\ldots$$
$$P(\sigma^{-1}(j) = k) \propto e^{-\sum_{i\notin\sigma^{-1}(1:j-1)} R_{ik}}. \quad (21)$$

**Proof of Lemma 3** This follows from Neal [2003].

**Proof of Lemma 5**

$$\begin{aligned} \sum_{s=0}^{n} \tilde{B}eta(s+a, b, n) & \\ &= \int_0^\infty \sum_{s=0}^{n} e^{-\theta(s+a)} \left(\frac{1-e^{-(n+1)\theta_j}}{1-e^{-\theta_j}}\right)^{-b+1} d\theta \\ &= \int_0^\infty e^{-\theta a} \frac{1-e^{-(n+1)\theta_j}}{1-e^{-\theta_j}} \left(\frac{1-e^{-(n+1)\theta_j}}{1-e^{-\theta_j}}\right)^{-b+1} d\theta \\ &= \tilde{B}eta(a, b-1, n). \end{aligned}$$

**Proof of Lemma 6** This follows by direct calculus.

**Proof of Lemma 7** The crucial observation here is the same as in Lemma 1: since $N = 1$, $L_\sigma(R_j(\pi)) = s_j(\pi \mid \sigma) = V_j(\sigma \mid \pi)$ by (1). As a consequence, the posterior of $\sigma$ is a product of multinomials, one for each $j = 1:t$:

$$P[V_j = v] \propto \tilde{B}eta(\nu r_j + v, \nu + 2, n - j). \quad (22)$$

We approximate $\tilde{B}eta(a, b, n)$ by $Beta(a, b)$. After $V_j$ is sampled, to construct $\sigma$ one places $\pi^{-1}(j)$ in the $V_j^{th}$ available position in $\sigma$. (See Meilă et al. [2007] for the detailed proof of this procedure.) The remaining $n-t$ positions are filled uniformly at random from the items not in $\pi$.

# References

C. E. Antoniak. Mixtures of Dirichlet processes with applications to Bayesian nonparametric problems. *Ann Stat*, 2(6):1152–1174, 1974.

D. M. Blei and M. I. Jordan. Variational inference for Dirichlet process mixtures. *Bayes Anal*, 1(1):121–144, 2006.

L. M. Busse, P. Orbanz, and J. Bühmann. Cluster analysis of heterogeneous rank data. In *Proceedings of ICML*, 2007.

M. A. Fligner and J. S. Verducci. Distance based ranking models. *J Roy Stat Soc B Met*, 48(3):359–369, 1986.

M. A. Fligner and J. S. Verducci. Multistage ranking models. *J Am Stat Assoc*, 83(403):892–901, 1988.

K. Goldberg, T. Roeder, D. Gupta, and C. Perkins. Eigentaste: A constant time collaborative filtering algorithm. *Inform Retrieval*, 4(2):133–151, 2001.

I. C. Gormley and T. B. Murphy. Analysis of Irish third-level college applications data. *J Roy Stat Soc A Sta*, 169(2):361–379, 2006.

I. C. Gormley and T. B. Murphy. Exploring voting blocs within the Irish electorate: a mixture modeling approach. *J Am Stat Assoc*, 103(483):1014–1027, 2008a.

I. C. Gormley and T. B. Murphy. A mixture of experts model for rank data with applications in election studies. *Ann Appl Stat*, 2(4):1452–1477, 2008b.

J. Guiver and E. Snelson. Bayesian estimation for Plackett-Luce ranking models. In *Proceedings of ICML*, 2009.

S. Jain and R. M. Neal. Splitting and merging components of a nonconjugate Dirichlet process mixture model. *Bayes Anal*, 2(3):445–472, 2007.

G. Lebanon and Y. Mao. Non-parametric modeling of partially ranked data. *J Mach Learn Res*, 9:2401–2429, 2008.

P. Liang, S. Petrov, M. I. Jordan, and D. Klein. The infinite PCFG using hierarchical Dirichlet processes. In *Proceedings of EMNLP*, 2007.

M. Meilă. Comparing clusterings—an information based distance. *J Multivariate Anal*, 98:873–895, 2007.

M. Meilă and L. Bao. Estimation and clustering with infinite rankings. In *Proceedings of UAI*, 2008.

M. Meilă, K. Phadnis, A. Patterson, and J. Bilmes. Consensus ranking under the exponential model. In *Proceedings of UAI*, 2007.

R. M. Neal. Markov chain sampling methods for Dirichlet process mixture models. *J Comput Graph Stat*, 9(2):249–265, 2000.

R. M. Neal. Slice sampling. *Ann Stat*, 31(3):705–767, 2003.

C. E. Rasmussen, B. J. de la Cruz, Z. Ghahramani, and D. L. Wild. Modeling and visualizing uncertainty in gene expression clusters using Dirichlet process mixtures. *IEEE/ACM T Comput Bi*, 6(4):615–628, 2009.

E. B. Sudderth, A. Torralba, W. T. Freeman, and A. S. Willsky. Describing visual scenes using transformed dirichlet processes. In *Advances in NIPS*, 2005.

Y. W. Teh, M. I. Jordan, M. J. Beal, and D. M. Blei. Hierarchical Dirichlet processes. *J Am Stat Assoc*, 101 (476):1566–1581, 2006.